\documentclass[journal]{IEEEtran}
\usepackage{graphicx}
\usepackage[table,xcdraw]{xcolor}
\usepackage{colortbl}
\usepackage{amssymb}
\usepackage{multirow}
\usepackage{CJKutf8}
\usepackage{amsmath}
\usepackage{booktabs}
\usepackage{multirow} 
\usepackage{xcolor}
\usepackage{makecell}
\usepackage{array}
\usepackage{booktabs} 
\usepackage{float}
\title{How to Build Robust, Scalable Models for GSV-Based Indicators in Neighborhood Research}

\author{%
Xiaoya Tang,%
Xiaohe Yue$^{\dagger}$,%
Heran Mane$^{\dagger}$,%
Dapeng Li,%
Quynh Nguyen,%
Tolga Tasdizen%
\thanks{Xiaoya Tang is with University of Utah, Scientific Computing and Imaging Institute, email: xiaoya.tang@utah.edu.}%
\thanks{Xiaohe Yue is with University of Maryland, email: xyue@umd.edu.}%
\thanks{Heran Mane is with University of Maryland, email: hmane@umd.edu.}%
\thanks{Dapeng Li is with University of Alabama, email: dli43@ua.edu.}%
\thanks{Quynh Nguyen is with National Institute of Nursing Research, National Institutes of Health, email: quynh.nguyen3@nih.gov.}%
\thanks{Tolga Tasdizen is with University of Utah, Department of Electrical and Computer Engineering, email: tolga.tasdizen@utah.edu.}%
\thanks{$^{\dagger}$These authors contributed equally to this work.}%
}

\begin{document}

\maketitle

{\begin{abstract}
A substantial body of health research demonstrates a strong link between neighborhood environments and health outcomes. Recently, there has been increasing interest in leveraging advances in computer vision to enable large-scale, systematic characterization of neighborhood
built environments. However, the generalizability of vision models across fundamentally different domains remains uncertain—for example, transferring knowledge from ImageNet to the distinct visual characteristics of Google Street View (GSV) imagery. In applied fields such as social health research, several critical questions arise: which models are most appropriate, whether to adopt unsupervised training strategies, what training scale is feasible under computational constraints, and how much such strategies benefit downstream performance. These decisions are often costly and require specialized expertise. 

In this paper, we answer these questions through empirical analysis and provide practical insights into how to select and adapt foundation models for datasets with limited size and labels, while leveraging larger, unlabeled datasets through unsupervised training. Our study includes comprehensive quantitative and visual analyses comparing model performance before and after unsupervised adaptation.
\end{abstract}

\section{Introduction}

\subsection{Built-in Environmental Charactertistics}
Neighborhood research has long been a cornerstone of social and health science, emphasizing how the characteristics of the places where people live and work influence physical and mental health\cite{burton2011communities}. Typically, a neighborhood covers physical design, social relationships, and environmental conditions\cite{patricios2002neighborhood}. Studies examine how physical, social and economic attributes of a local area contribute to health differences at the population level\cite{sirgy2002neighborhood}. Common domains include walkability, transportation networks, environmental pollution, land use, access to resources (food, gym, hospital, etc.), social connections, equity issues, safety level, cost of living in the community, and home values\cite{diez2010neighborhoods,sirgy2002neighborhood}. Neighborhood studies also investigate how neighborhood disrepair and disinvestment affect health outcomes, such as breast cancer\cite{plascak2022associations}. Through a multidimensional lens, neighborhood studies underscore that health is affected not only by individual habits but also by the broader systems that shape daily living conditions.

Over the past decade, methodologies in neighborhood research have evolved substantially. Early studies relied primarily on surveys and administrative data, which, while valuable, are costly, slow, and limited in their ability to capture the multidimensional characteristics of diverse communities\cite{yue2022using}. The emergence of geospatial technologies, high-resolution imagery, social media, drones, webcams, and other tools has transformed the field\cite{schootman2016emerging}. Moreover, advances in machine learning, computer vision, and spatial analytics now allow researchers to directly observe built-environment features, quantify visual cues of disorder, and detect neighborhood change over time. These developments have improved the precision and scalability of neighborhood measurement, enabling more detailed and accurate examinations of how environments affect residents' health across spatial and temporal contexts.

A study\cite{nguyen2016building} by Nguyen et al. used social media data to characterize neighborhood well-being and health behaviors. Using more than 80 million geotagged U.S.-based tweets, the researchers applied machine learning and spatial mapping methods to generate neighborhood-level indicators of happiness, diet, and physical activity. These social media–derived measures were correlated with socioeconomic disadvantage, urbanicity, and chronic disease prevalence, demonstrating that social media data can complement traditional neighborhood measures.

Building on these findings, Nguyen et al.\cite{nguyen2019using} developed a framework using more than 16 million Google Street View images across the United States to automatically characterize built-environment features such as road types, infrastructure density, and visible signs of development. By applying computer vision models to these street view images and linking them with county- and census-tract-level health data, their analyses revealed that areas with limited infrastructure have higher rates of obesity and diabetes, whereas more developed areas show more favorable health outcomes.

In another cross-sectional study of fatal collisions across the United States, researchers evaluated links between built-environment characteristics and collision prevalence. Features such as single-lane roads and street greenness at the census-tract level were associated with reduced prevalence of pedestrian and cyclist collisions\cite{nguyen2025leveraging}.  These studies demonstrate how large-scale street view imagery can serve as a powerful resource for monitoring built environments and understanding spatial health disparities.

When situating individuals within their surrounding environmental contexts to examine how neighborhood conditions interact with various factors in shaping health outcomes, contemporary studies increasingly recognize neighborhoods as dynamic systems rather than static settings. This perspective allows researchers to identify determinants of inequity and design effective, targeted interventions\cite{mallach2023dynamic}. Integrating neighborhood research with public health research can help inform policies and shape community strategies to improve health across diverse populations.  

\begin{figure*}[t]
  \centering
  \includegraphics[width=0.98\textwidth, keepaspectratio=false, height=0.55\textwidth]{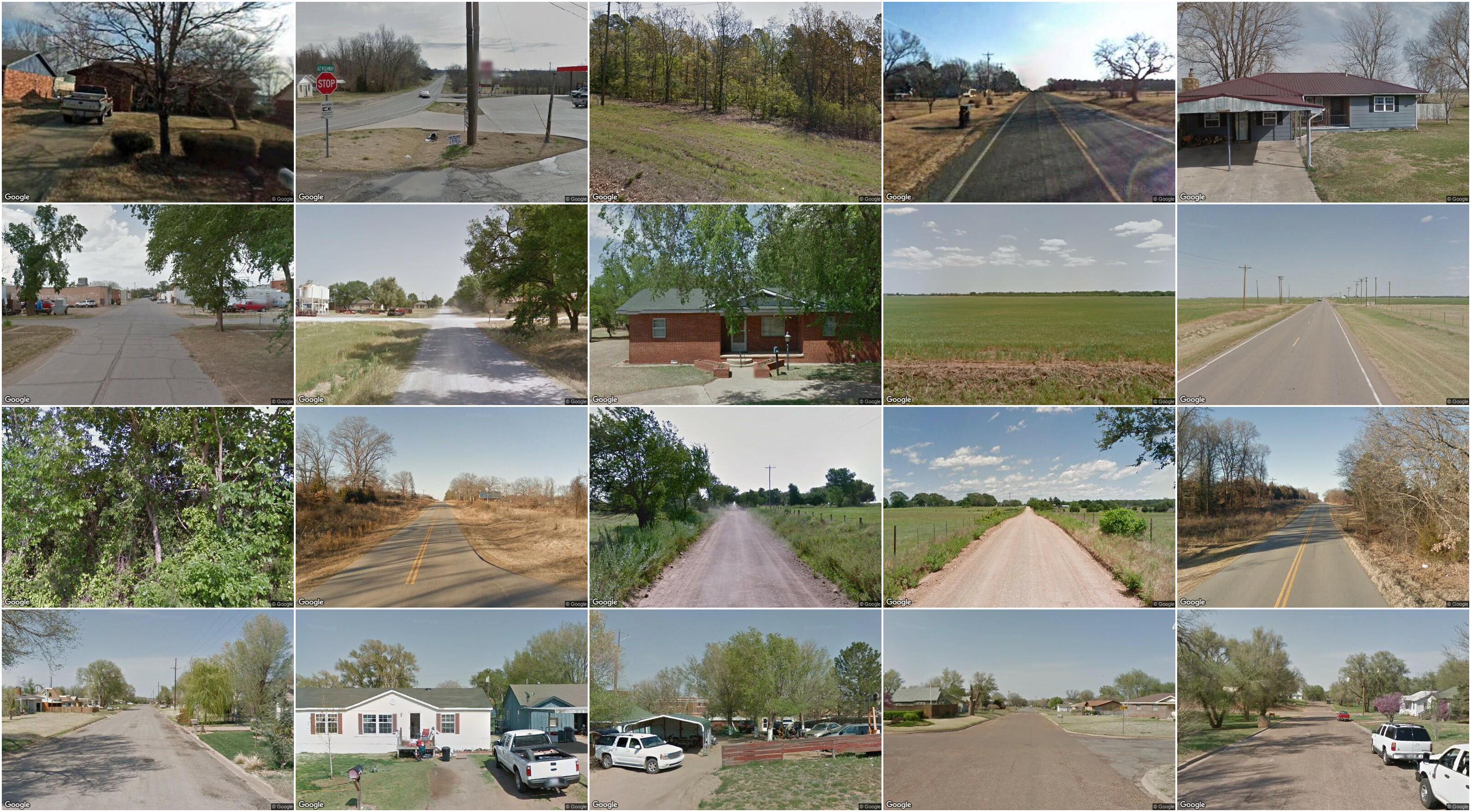}
  \caption{Random GSV image samples used in unsupervised training. Original image size: 640 x 440.}
  \label{fig:GSV samples}
\end{figure*}

Earlier neighborhood studies often relied on convolutional neural networks (CNNs) such as VGG and ResNet to recognize environmental features and derive neighborhood indicators. With the emergence of foundation models, researchers have increasingly turned to modern architectures such as Vision Transformers (ViTs), which are pretrained on large-scale imagery datasets like ImageNet (IN). ViTs offer advantages not only through stronger representation learning enabled by extensive pretraining but also through more interpretable attention maps, allowing researchers to gain insight into the features the model has learned rather than treating it as a black box. Using the Google Street View (GSV) dataset shown in Figure \ref{fig:GSV samples}, we evaluate several model architectures for neighborhood classification. The results show that contemporary architectures such as ViT and the more recent Vision Mamba (Vim) demonstrate strong potential when pretrained at scale. However, despite the strong performance of IN-pretrained models, a domain gap persists between ImageNet imagery and GSV images. To mitigate this discrepancy, we applied an unsupervised method to adapt these models without requiring annotations.

\subsection{Modern Deep Neural Networks}\label{section:Models}
\subsubsection{Transformer Architectures}
The Vision Transformer (ViT)\cite{dosovitskiy2020image}(Figure~\ref {fig:Architectures1}) has become a leading architecture across a wide range of language and computer vision benchmarks since its emergence, following the success of the self-attention mechanism\cite{vaswani2017attention}. It has now become a fundamental building block of Computer Vision (CV) models, Vision-Language Models (VLMs)\cite{zhang2024vision}, and Large Language Models (LLMs)\cite{zhao2023survey}. ViTs have demonstrated a higher performance ceiling than traditional convolutional neural networks (CNNs)\cite{krizhevsky2012imagenet}, such as ResNets\cite{he2016deep}, particularly after large-scale pretraining or when provided with sufficient data to acquire the necessary inductive biases.

However, the lack of inductive bias has been a persistent challenge for Transformer-based models—especially in the vision domain—due to the modality shift from language to images while still relying on the same patch tokenization process. In a vanilla ViT\cite{dosovitskiy2020image}, the general workflow includes patch tokenization/embedding, stacked attention blocks, and a task-specific head. An image is uniformly divided into fixed-size patches (typically 16×16), which are then projected via a convolutional layer into a sequence of vector representations, or tokens. This token sequence is subsequently processed through a series of attention layers. The self-attention mechanism, shown in Eq.~\ref{eq:multihead self-attention}, computes relationships among all tokens within the sequence, yielding a globally contextualized representation of the entire image. For an input sequence $\mathbf{x}\in \mathbb{R}^{N\times D}$, query, key, and value matrices $(\mathbf{q},\mathbf{k},\mathbf{v})$ are obtained via learned linear projections. The attention weights $A_{ij}$ quantify the similarity between the $i$-th query and the $j$-th key, and are used to produce a weighted combination of all value vectors. Multi-Head Self-Attention (MSA) extends this operation by performing $n_h$ independent attention computations (“heads”) in parallel and projecting the concatenated outputs back to the feature dimension, allowing the model to capture diverse relational patterns across tokens.
\textbf{\begin{equation}
    \begin{aligned}
\left[\mathbf{q},\mathbf{k},\mathbf{v}\right]&=\mathbf{x}\mathbf{W}_{qkv},\mathbf{W}_{qkv}\in \mathbb{R}^{D\times 3D_h},D_h=\frac{D}{n_h},\\
        A&=\text{Softmax}\left( \frac{\mathbf{q}\mathbf{k}^T}{\sqrt{D_h}}\right),A \in \mathbb{R}^{N\times N},\\
        \text{SA}(\mathbf{x})&=A\mathbf{v}, \\
        \text{MSA}(\mathbf{x})&=\left[\text{SA}_1(\mathbf{x});\text{SA}_2(\mathbf{x});...;\text{SA}_{n_h}(\mathbf{x})\right]\mathbf{W}_{msa},\\
        \mathbf{W}_{msa}&\in \mathbb{R}^{n_h\cdot D_h\times D}.
    \end{aligned}
    \label{eq:multihead self-attention}
\end{equation}}
ViT-based architectures have exhibited strong holistic understanding for classification tasks and fine-grained comprehension for dense prediction tasks such as segmentation, across diverse data modalities including natural images\cite{dosovitskiy2020image}, videos\cite{fan2021multiscale}, audios\cite{li2025audio}, and medical data encompassing both imaging and temporal modalities\cite{tang2024hierarchical,tang2025dynamic}. Although ViTs are generally more computationally expensive than CNNs of comparable size, they are believed to possess greater representational flexibility and scalability across multiple tasks, provided there are sufficient inductive biases, either from the data or from architectural design choices.

To mitigate the lack of inductive bias—particularly translation equivariance and spatial locality \cite{lee2021vision,raghu2021vision}—which are naturally encoded in CNNs, a large body of work has explored methods for integrating such priors into ViTs without relying solely on large training data. One research direction introduces auxiliary self-supervised objectives to make better use of limited data. Recently, Das et al. \cite{das2024limited} conducted a thorough study on the training schemes of Self-Supervised Learning (SSL) tasks for ViTs and found that jointly optimizing ViTs for the primary task (fine-tuning on limited labeled data) and a Self-Supervised Auxiliary Task (SSAT) is more effective than performing SSL and fine-tuning sequentially. Another direction involves injecting multi-scale inductive biases from CNNs into ViTs by integrating convolutional operations at various stages: prior to patch embedding, within attention blocks\cite{dai2021coatnet}(e.g., replacing linear projections as in \cite{wu2021cvt}), via pooling mechanisms\cite{fan2021multiscale}, or through hybrid hierarchical designs\cite{tang2024duoformer,wang2023crossformer++}.
Subsequent advancements have explored architectures with hierarchical downsampling\cite{wang2021pyramid,wang2022pvt,li2022mvitv2} or intricate windowing strategies\cite{liu2021swin,dong2022cswin}, addressing the limitations of uniform, single-scale patch grids. 
Concurrently, significant effort has gone into enhancing the locality of ViTs while improving the computational efficiency. This has led to the development of local attention\cite{parmar2018image}, an early form of windowed attention in which self-attention is confined to spatially local regions, further enabling pixel-level detail modeling \cite{hassani2023neighborhood}. Moreover, by incorporating deformable convolutions or other dynamic mechanisms, models can support deformable or adaptive attention within windows\cite{pan2023slide}, effectively combining the representational flexibility of attention with the efficiency and inductive biases of convolutions. 

\subsubsection{Efficient Attention}
Improving computational efficiency has become one of the central focuses in the evolution of Transformer–based architectures. While Transformers excel at modeling global dependencies through a fully connected self-attention graph, they inherently suffer from the notorious $O(N^2)$ computational and memory complexity, where N denotes the number of nodes(tokens, or patches) in the fully connected graph. Linear Attention \cite{katharopoulos2020transformers} reformulates the self-attention mechanism by expressing the softmax similarity function as a dot product between kernel feature maps, leveraging the associativity of matrix multiplication to achieve a linear computational complexity with respect to sequence length(N). This formulation also reveals the intrinsic connection between Transformers and recurrent neural networks (RNNs), enabling efficient autoregressive modeling with faster training and inference. Window-based attentions are able to alleviate the same issue from a different perspective, by restricting self-attention to local regions. When the window or kernel size is relatively small compared to the input resolution, the computational complexity can be significantly reduced \cite{liu2021swin}, as demonstrated in local attention mechanisms for vision data\cite{hassani2023neighborhood}.
Building upon this, Dilated Neighborhood Attention Transformer(DiNAT) \cite{hassani2022dilated} extends Neighborhood Attention \cite{hassani2023neighborhood} into a dilated local attention capable of preserving locality, maintaining translational equivariance, and exponentially expanding the receptive field.

Different from window attention, which is a reduced form of patch-to-patch attention, PaCa-ViT\cite{grainger2023paca} learns to cluster patches into a predefined number of $M$ clusters that serve as keys and values, relaxing the quadratic complexity to linear and demonstrating greater efficiency than PVT models \cite{wang2021pyramid} and Swin \cite{liu2021swin}.

To further lower both time and space complexity, researchers have explored token pruning methods for various downstream tasks such as segmentation\cite{tang2023dynamic}. In language tasks involving long-context inputs, where the number of tokens can reach millions, additional studies have leveraged the sparsity of attention weights in pretrained LLMs. These include heuristic pruning based on attention patterns \cite{jiang2024minference}, low-rank decompositions \cite{wang2020linformer}, learned thresholding \cite{kim2022learned}, predictive pruning of token indices \cite{akhauri2025tokenbutler}, and entropy-inspired approaches \cite{xiong2024uncomp} that dynamically allocate computation based on content relevance.\\
\begin{figure*}[t]
  \centering
  \includegraphics[width=\textwidth]{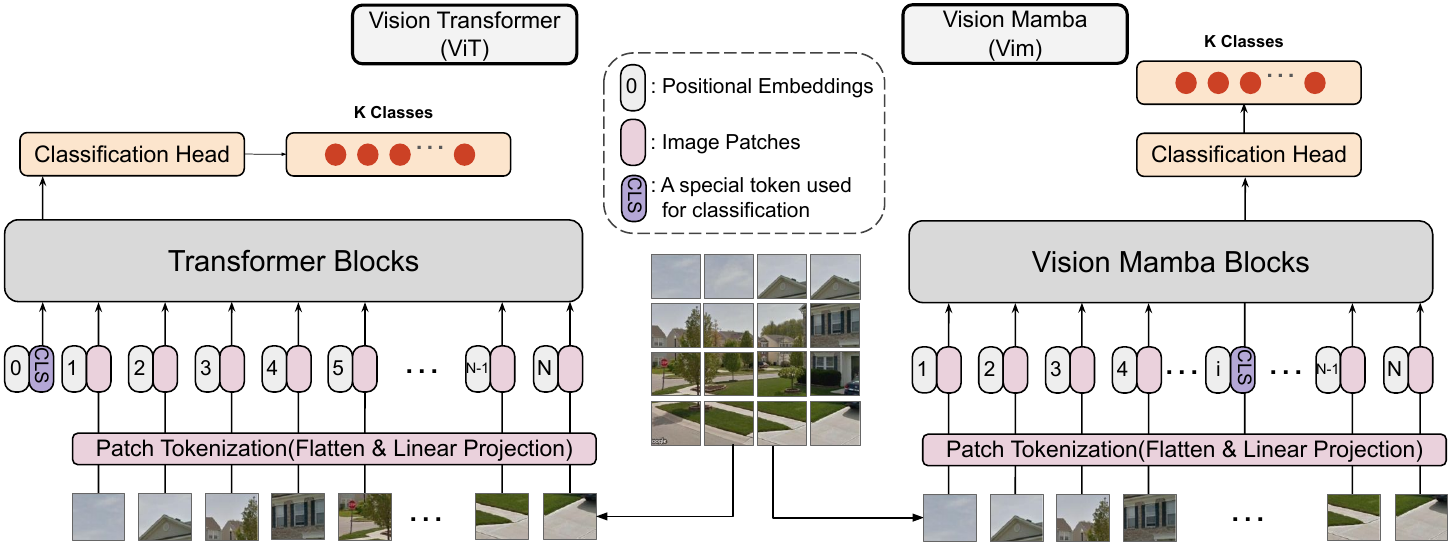}
  \caption{Model Architectures(Left: ViT; Right: Vim). The gray rectangles labeled from 0 to N represent the positional embeddings. }
  \label{fig:Architectures1}
\end{figure*}

\subsubsection{State Space Model}\label{section:SSM}
Despite the existing efforts to improve the efficiency of attention computation while maintaining the representational capacity of ViTs, a growing body of research has been exploring alternative architectures as potential successors to modern neural networks. Among these, Mamba \cite{gu2024mamba} revisits state-space models (SSMs) and demonstrates strong performance on 1D long-sequence modeling with linear or near-linear complexity with respect to input length. This has inspired work on constructing general-purpose vision backbones built entirely on SSMs. The key challenges in applying SSMs to vision are: (1) encoding the position-sensitive nature of 2D spatial structures and (2) capturing global visual context, since the original 1D Mamba is unidirectional and lacks inherent positional awareness.

SSMs originate from continuous-time dynamical systems, mapping an input sequence $u(t)\in \mathbb{R}^L$ to an output $y(t)\in \mathbb{R}^L$ through a hidden state $h(t) \in \mathbb{R}^N$, where $t$ denotes time. A continuous-time linear SSM is expressed as the Ordinary Differential Equation(ODE) in Eq. \ref{eq:mamba ODE}, where $\mathbf{A}\in \mathbb{R}^{N\times N}$ is the state transition matrix. $\mathbf{B}\in \mathbb{R}^{N\times 1}$, and $\mathbf{C}\in \mathbb{R}^{1\times N}$ are the input and output projection matrices, respectively. 
\begin{equation}
\begin{aligned}
h'(t)&=\mathbf{A}h(t)+\mathbf{B}u(t),\\
y(t)&=\mathbf{C}h(t).
\end{aligned}
\label{eq:mamba ODE}
\end{equation}
To integrate SSMs into deep learning, this continuous model must be discretized. A standard approach is the Zero-Order Hold (ZOH) method, which yields discrete matrices $\bar{\mathbf{A}}$ and $\bar{\mathbf{B}}$, parameterized by a times step $\mathbf{\Delta}$, as shown in Eq. \ref{eq:mamba discretization}. 
\begin{equation}
\begin{aligned}
\bar{\mathbf{A}} &= \exp(\Delta\mathbf{A}),\\
\bar{\mathbf{B}} &= (\Delta\mathbf{A})^{-1}(\exp(\Delta\mathbf{A})-\mathbf{I})\cdot \Delta\mathbf{B}.
\end{aligned}
\label{eq:mamba discretization}
\end{equation}
This produces the discrete recurrence in Eq.~\ref{eq:mamba discretization-2}:
\begin{equation}
\begin{aligned}
h_t&=\bar{\mathbf{A}}h_{t-1}+\bar{\mathbf{B}}u_t,\\
y_t&=\mathbf{C}h_t.
\end{aligned}
\label{eq:mamba discretization-2}
\end{equation}
S4 \cite{gu2021efficiently} and Mamba (also known as S6)\cite{gu2024mamba} both adopt this discretized SSM form above. However, Mamba introduces a selective input-dependent parameterization, where $\mathbf{\Delta}$, $\mathbf{B}$ and $\mathbf{C}$ are conditioned on the input, while $\mathbf{A}$ is learned as a global parameter. In practice, for input $u \in \mathbb{R}^{b\times L\times N}$, Mamba produces $\mathbf{\Delta}\in \mathbb{R}^{b\times L\times D}$, $\mathbf{B,C}\in \mathbb{R}^{b\times L\times N}$, dynamically through the Selective Scan mechanism. Here, $b$ represents the batch size, $L$ denotes the sequence length, $D$ is the feature dimension, and $N$ can be viewd as the hidden size of SSM, e.g., 16. Then $\mathbf{\Delta}$ and $\mathbf{B}$ are used to transform the $\bar{\mathbf{A}}$ and $\bar{\mathbf{B}}$.

Zhu et al.\cite{zhu2024vision} adapt S4 and Mamba to vision in Vision Mamba (Vim). Images are partitioned into patches, projected to tokens, and processed as sequences—similar to ViT. To account for the non-causal nature of images, Vim introduces bidirectional selective scanning and position embeddings, enabling global context modeling and spatial awareness. The model overview is shown in Fig.~\ref{fig:Architectures1}. A 2D image $\mathbf{t} \in \mathbb{R}^{H \times W \times C}$ is first divided into non-overlapping patches of size $P \times P$, which are then flattened into vectors, forming $\mathbf{x_p} \in \mathbb{R}^{J \times (P^2C)}$, where $(H, W)$ is the spatial resolution of the input, $C$ is the number of channels, and $J$ is the number of patches. Each patch vector is then linearly projected into a $D$-dimensional embedding using a learnable projection matrix $\mathbf{W}\in \mathbb{R}^{(P^2C)\times D}$, and a positional embedding $\mathbf{E}_{pos}\in \mathbb{R}^{(J+1)\times D}$ is added, as shown in Eq.~\ref{eq:positinal embeddings}. Following the ViT design, a learnable [CLS] token $\mathbf{t}{cls}$ is introduced to summarize the global representation of the entire patch sequence. Based on ablation studies, $\mathbf{t}{cls}$ is inserted in the middle of the sequence rather than at the beginning. As in ViT, the entire sequence is then used as the input of a stack of SSM blocks in Vim encoder, where sequence $\mathbf{T}_0$ was processed from the forward and backward directions. The output class token $\mathbf{t}{cls}$ after the final layer of Vim block is used for prediction. 
\begin{equation}
    \begin{aligned}
        \mathbf{T}_0 &= [\mathbf{t}_{cls};\mathbf{t}_{p}^1\mathbf{W};\mathbf{t}_{p}^2\mathbf{W};...;\mathbf{t}_{p}^J\mathbf{W}]+\mathbf{E}_{pos}]\\[4pt]
    \end{aligned}
    \label{eq:positinal embeddings}
\end{equation}
Almost concurrently, Liu et al.~\cite{liu2024vmamba} integrated visual state-space (VSS) blocks into a hierarchical backbone. In their design, the 2D Selective Scan (SS2D) module performs four-way (cross) scanning and merging to traverse the spatial domain. Specifically, SS2D unfolds feature maps into sequences along four different traversal paths. Each sequence is then processed in parallel by an independent S6 block, and the resulting sequences are reshaped and merged back into a 2D feature map. By leveraging these complementary 1D traversal directions, SS2D enables each pixel to aggregate information from all other spatial positions, effectively establishing a global receptive field in 2D space. Through this mechanism, VMamba~\cite{liu2024vmamba} adapts S6 to visual data without sacrificing key advantages of self-attention, namely global context modeling and dynamic, content-dependent weighting (in contrast to CNNs that use fixed convolutional kernels)~\cite{han2021connection}. The comparison of different architectures is shown in Figure~\ref{fig:receptive field}.

Several subsequent works on visual Mamba architectures primarily differ in their scanning strategies. The raster scan is the most widely adopted in current implementations\cite{xu2024visual}. More recently, Liu et al.~\cite{liu2025defmamba} introduced a deformable scanning approach to reduce the loss of structural information brought by fixed scanning paths.

On ImageNet classification, many Visual Mamba architectures outperform the CNN-based MambaOut~\cite{yu2025mambaout} and hierarchical Transformer baselines such as Swin~\cite{liu2021swin}. Visual Mambas have also demonstrated strong performance across dense prediction tasks—including object detection and semantic segmentation~\cite{xu2024visual}—and have been extended to a wide range of modalities, such as X-ray images~\cite{yang2024cardiovascular}, digital pathology~\cite{nasiri2024vim4path}, 3D medical images~\cite{tsai2024uu}, remote sensing~\cite{chen2024rsmamba}, and video understanding~\cite{li2024videomamba}. These applications benefit from Mamba’s ability to model long-range spatio-temporal dependencies efficiently.

However, as observed by Xu et al.~\cite{xu2024visual}, Visual Mamba models demonstrate competitive performance, yet there are still cases where they do not fully match the most advanced Transformer architectures. Additionally, most current implementations are relatively small in scale, and how to scale Mamba-based models efficiently continues to be an active direction for future research. Similar to ViT, pure Mamba architectures also struggle to capture local fine-grained details, which are crucial for low-level visual tasks. To address this limitation, recent works incorporate convolutions, channel attention mechanisms~\cite{guo2024mambair}, and multi-scale or hybrid scanning strategies~\cite{shi2024multi}.
\begin{figure*}[t]
  \centering
  \includegraphics[width=\textwidth]{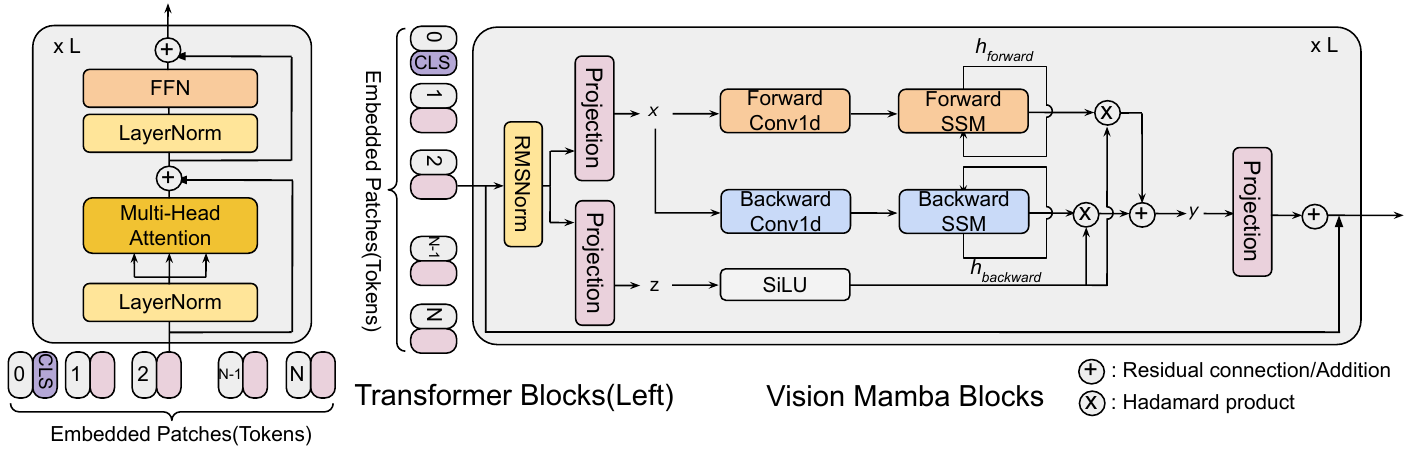}
  \caption{Model Architectures(Left: ViT; Right: Vim). The gray rectangles labeled from 0 to N represent the positional embeddings. }
  \label{fig:Architectures2}
\end{figure*}
\begin{figure}[t]
  \centering
  \includegraphics[width=0.42\textwidth]{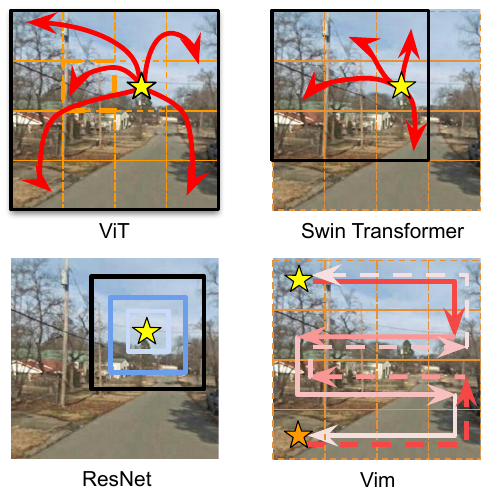}
  \caption{Comparison of token dependencies(Lines with arrows) and receptive field(Black boxes) for different architectures. }
  \label{fig:receptive field}
\end{figure}

\subsection{Unsupervised Techniques}\label{sec:unsupervised}
Besides incorporating appropriate inductive biases through architectural design, researchers have also leveraged unsupervised techniques in modern deep neural networks, such as Transformers, solving the problem of limited amount of labeled data by exploiting unlabeled data. A major line of research in unsupervised and self-supervised learning is contrastive learning, which encourages representations from the same image (positive pairs) to be close in the latent space while pushing apart representations from different images (negative pairs). In this section, we focus on self-supervised methods trained purely on visual data and exclude multimodal approaches such as CLIP \cite{radford2021learning}.

SimCLR v1 \cite{chen2020simple}, a representative work in contrastive learning, systematically analyzed the key factors contributing to its success. It constructs positive pairs by applying diverse data augmentations to an image, and employs a lightweight projection head to map features into a latent space where a contrastive loss is applied. The study found that the combination of random cropping and color distortion is crucial for strong performance. Subsequent work further demonstrated that learning from multiple augmented views leads to representations that better capture the underlying scene semantics \cite{tian2020contrastive}. However, SimCLR largely benefits from large batch sizes, which provide sufficient negative samples. To alleviate this limitation, MoCo v1 \cite{he2020momentum} introduced a momentum encoder and a dynamic memory queue that maintains a large set of negative keys, effectively decoupling the dictionary size from the mini-batch size. This allows contrastive learning with smaller batches while retaining negative sample diversity. MoCo v2 \cite{chen2020improved} further improved upon SimCLR’s findings by incorporating an MLP projection head and stronger data augmentations, achieving competitive results with more efficient training.

Beyond contrastive paradigms, other self-supervised approaches have demonstrated strong representation learning capabilities. A prominent example is masked image modeling (MIM), a recent and influential paradigm for visual pre-training \cite{bao2021beit}. Masked Autoencoders (MAE) \cite{he2022masked} adopt this idea by randomly masking a large proportion of image patches (around 75\%) before encoding and reconstructing the missing pixels via a lightweight decoder. This process compels the encoder to learn semantically meaningful features. After pretraining, the decoder is discarded, and the encoder is applied directly to downstream tasks. Another line of work addresses the batch-size sensitivity and reliance on negative samples, exemplified by BYOL \cite{grill2020bootstrap}. BYOL trains an online network to predict the representation of a target network fed with a differently augmented view of the same image, eliminating the need for negative samples and contrastive loss. This design achieves stable training even with small batches and is less sensitive to augmentation choices compared to contrastive methods.

At the era of Transformer architectures, self-supervised methods have been adapted and applied to Vision Transformers (ViTs). MoCo v3 \cite{chen2021empirical}, an incremental improvement over MoCo v1\cite{he2020momentum} and MoCo v2\cite{chen2020improved}, investigated the instability of ViTs in self-supervised learning. It found that adding a random patch projection layer slightly stabilized training. iBOT\cite{zhou2021ibot} performs masked prediction using a teacher network as an online tokenizer. Two losses are minimized in this approach: one for self-distillation between cross-view [CLS] tokens, and another for self-distillation between in-view patch tokens, where some masked tokens in the student network are reconstructed using the teacher network’s outputs as supervision. The two objectives are summed without scaling.

DINO \cite{caron2021emerging} is a widely used self-distillation approach that requires no labels. It also employs a teacher-student architecture and has proven particularly effective for ViTs. In DINO, an image $x$ is randomly cropped into two global views $x_1^g$  and $x_2^g$ and several local views of smaller spatial extent. All crops are processed by the student network, while only the global views are fed into the teacher. The core idea is to train a smaller student network $g_{\theta_s}$ to replicate the behavior of a larger teacher network $g_{\theta_t}$  by leveraging its output probability distributions, which convey richer information than conventional hard class labels. Their output probability distributions over $K$ dimensions are denoted by $P_s$ and $P_t$, obtained by normalizing the output of the network $g$ with a softmax function, given in Eq. \ref{eq:DINO output prob}, with temperature parameters $\tau_s,_t>0$ controlling the sharpnesses of the output distributions. 
\begin{equation}
    \begin{aligned}
        P_s(x)^i =  \frac{\exp(g_{\theta_s}(x)^i/\tau_s)}{\sum_{k=1}^K \exp(g_{\theta_s}(x)^k/\tau_s)}
    \end{aligned}
    \label{eq:DINO output prob}
\end{equation}
\begin{equation}
    \begin{aligned}
        \min_{\theta_s}\sum_{x\in\{x_1^g,x_2^g\}}&\sum_{\substack{x' \in V \\ x' \neq x}}H(P_t(x),P_s(x')),\\
        H(a,b)&=-a\log b.
    \end{aligned}
    \label{eq:DINO loss}
\end{equation}
To stabilize training, the teacher outputs are centered over the mean and softened using temperature scaling. By minimizing the similarity loss between teacher and student representations given in Eq. \ref{eq:DINO loss}, in which V is a set of different views, containing global and local ones. DINO enforces local-to-global correspondence. During training, the student network parameters are updated via gradient backpropagation, whereas the teacher network parameters are updated through an exponential moving average. Self-supervised ViT features learned via DINO have been shown to capture explicit semantic information useful for diverse downstream tasks, exhibiting strong transferability. DINO v2 \cite{oquab2023dinov2} scaled this self-distillation paradigm to massive model sizes (~billions of parameters) and extremely large batch sizes (around 65k). DINO v3 \cite{simeoni2025dinov3}, published at the time of writing, further scales DINO to billions of parameters and images. It addresses the problem of dense feature degradation—where patch-level representations collapse into similar embeddings over long training runs—by introducing Gram Anchoring, a regularization technique that encourages the student model’s Gram matrix (pairwise patch similarities) to remain close to that of a more stable earlier teacher network. This strategy has been shown to repair degraded local features after approximately one million iterations.

\section{Model Choices}
In this section, we describe the model architectures and sizes evaluated in our study and explain the rationale behind these choices.

Scaling laws \cite{kaplan2020scaling} have been extensively studied in language models and, more recently, in discriminative vision models \cite{zhai2022scaling}, revealing predictable power-law relationships among model size, dataset size, compute budget, and performance. Scaling models and data can sometimes lead to qualitatively new behaviors, often referred to as emergent abilities. In the vision domain, similar effects have been reported, although they are less formally characterized. For example, DINO \cite{caron2021emerging} observes representation phase transitions under self-supervised training, and SAM \cite{kirillov2023segment} demonstrates strong zero-shot generalization to unseen categories, which can be viewed as an emergent-like property arising from large-scale training and diverse data. These capabilities generally strengthen with increased model capacity, particularly for larger architectures such as ViT-B/L/H or even ViT-G with billions of parameters. However, training and fine-tuning models at this scale are computationally expensive, and the resulting performance gains can be modest \cite{zhai2022scaling}, limiting their practical applicability.

Despite observations that jointly scaling compute, model size, and data improves representation quality on ImageNet and other well-known benchmarks \cite{zhai2022scaling}, smaller models can remain surprisingly competitive\cite{nasiri2024vim4path}. For instance, ViT-S/16 achieves performance comparable to ViT-B/32 under the ImageNet-21k linear 10-shot evaluation protocol. Nonetheless, the generalizability of smaller models to substantially different tasks—such as transferring from ImageNet to environmental characteristics in GSV imagery—remains uncertain. In applied settings like social health research, key questions persist: which models to choose, whether to post-train or fine-tune, and at what scale such training should be applied, considering computational cost constraints. These practices are often expensive and require expert knowledge. In this work, we aim to provide practical insights by exploring these questions.

Hierarchical transformers are often more suitable for tasks requiring dense, fine-grained features—such as medical image analysis or segmentation, where pixel- or region-level understanding is crucial. However, the semantic patterns in our GSV data are typically not fine-grained (i.e., built-in environmental indicators such as streetlights, single-family houses, etc.). Therefore, employing hierarchical architectures such as the Swin Transformer(Swin)\cite{liu2021swin} and DiNAT \cite{hassani2022dilated} may not be urgent or necessary. Simultaneously, Mamba shows great potential as a more recent visual foundation architecture, which we also aim to compare experimentally with vanilla architectures on our neighborhood research tasks. Consequently, we fixed our chosen architectures to ViT and Vim, while using ResNet and Swin as baselines representing CNNs and hierarchical Transformers, respectively, to provide a clear comparison across different architectures on GSV imagery. We discuss the theoretical details of these models in Section~\ref{section:Models} and illustrate the token dependencies in different types of models in Figure \ref{fig:receptive field}.

For model sizes, we selected ViT-S and ViT-B as representatives of smaller and medium-sized ViT models, and Vim-S and Vim-B as representatives of smaller and medium-sized Vision Mamba models. This allows us to evaluate the practical potential of the two types of foundational vision models in social health research applications while balancing performance with computational cost. We also aim to examine how model size influences performance on GSV imagery. We offer the details of models used in the paper in Table \ref{tab:model variants}.

For model initialization, we post-trained ViT and Vim from ImageNet-pretrained weights. Specifically, we used ViT unsupervisedly pretrained on ImageNet via DINO \cite{caron2021emerging} and Vim supervisedly pretrained on ImageNet \cite{zhu2024vision}. Beyond comparing architectures and model sizes, we are also interested in the performance differences between models pretrained with labels versus self-supervised learning (SSL) techniques. 
\begin{table*}[]
    \centering
    \begin{tabular}{cccccccc}
    \toprule
    { Model} & { Depth} & { Hidden size} & { MLP size}& {Attention Heads} & {Params.(M)}& {ImageNet acc@1} & {ImageNet acc@5}   \\ \midrule
    {ResNet50}   & {1,3,5,5}       & {128/256/512/1024 }       & { -} & { -}  & { 23.51} & {80.86} & { 95.43}        \\ \specialrule{.03em}{0.3ex}{0.3ex}
    { Swin-S} & {2,2,18,2}   & {96,192,384,768}    & { 384,768,1536,3072}  &{3,6,12,24}& { 48.84} & {83.20}    & { 96.36}     \\ \specialrule{.03em}{0.3ex}{0.3ex}
    { ViT-S}  & { 12}    & { 384}  & {1536}    & {6} &{ 21.67} &{77.0}&{-}  \\ \specialrule{.03em}{0.3ex}{0.3ex}
    { Vim-S} & { 24}    & { 384}  & { -}    & { -} &{ 25.62} &{81.6}&	{95.4}    \\ \specialrule{.03em}{0.3ex}{0.3ex}
    { Swin-B} & {2,2,18,2}   & {128,256,512,1024}    & {512,1024,2048,4096}  &{4,8,16,32}& { 86.75} & {83.58}    & { 96.64}  \\ \specialrule{.03em}{0.03em}{0.03em}
    { ViT-B} & { 12}   & { 768}    & { 3072}  & { 12}    & {85.81}   &{80.1}&{-} \\ \specialrule{.03em}{0.3ex}{0.3ex}
    { Vim-B} & { 24}   & { 768}    & {-}  & {-}    & {96.83} &{81.9}&	{95.8}     \\ \bottomrule
    \end{tabular}
    \caption{Details of Model variants, weights pretrained on ImageNet1k were loaded. A list of depths and hidden sizes indicates a hierarchical or pyramid model structure. Accuracies on ImageNet are from corresponding paper or official document.}
    \label{tab:model variants}
\end{table*}

\section{Unsupervised Training}\label{sec:unsupervised training}
\begin{figure*}[t]
  \centering
  \includegraphics[width=\textwidth, keepaspectratio=false, height=0.44\textwidth]{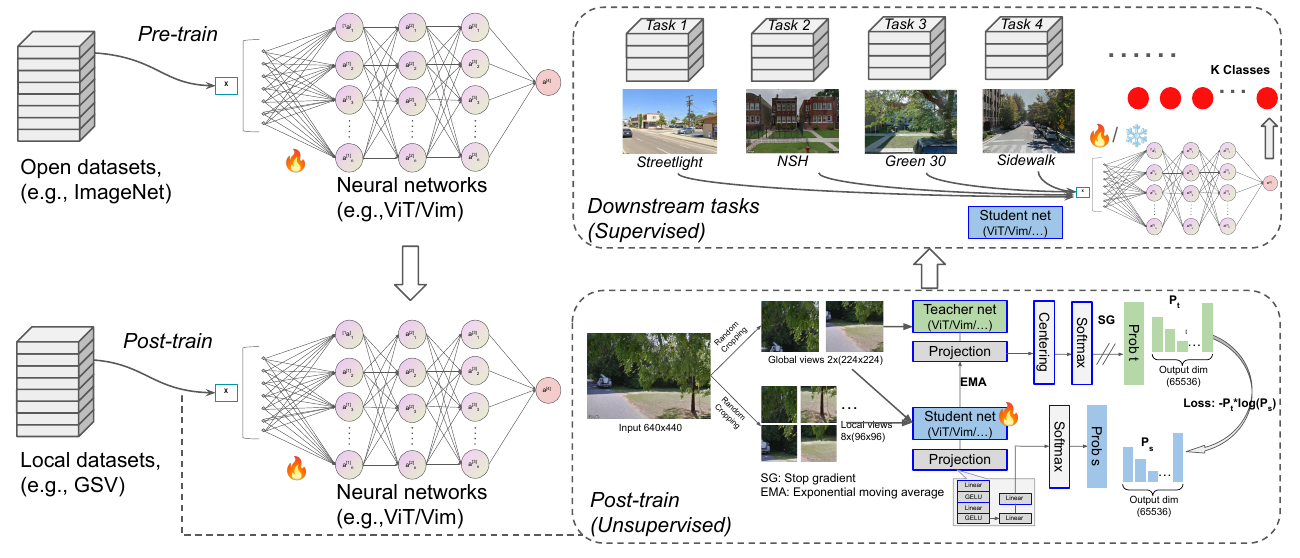}
  \caption{Illustration of our post-training and evaluation pipeline. The backbone network can be instantiated with any architecture; here we use ViT and Vim. During post-training, model weights are updated(indicated by “fire”). During inference, we either fine-tune or freeze the pretrained backbone(indicated by “freeze”) while updating the classification head.}
  \label{fig:training framework}
\end{figure*}
To learn strong and transferable representations from massive, unlabeled GSV imagery, we adopt SSL techniques, as introduced in Section \ref{sec:unsupervised}. GSV images frequently contain visually similar or semantically ambiguous patterns—often even between geographically distant locations or, conversely, between adjacent viewpoints—as illustrated in Figure \ref{fig:GSV samples}. Elements such as vegetation, roadway surfaces, sky appearance, or common architectural structures may look nearly identical across samples, while regional style variations can remain subtle.

Given this characteristic of the data, we opt for knowledge-distillation–based SSL rather than contrastive learning, since contrastive methods require negative samples that must be semantically dissimilar and explicitly pushed apart in the representation space. This requirement does not hold well for GSV imagery. DINO~\cite{caron2021emerging} has proven to be one of the most effective distillation-based approaches for training ViT models, and thus we choose it to post-train both a ViT and a Vim on GSV data. A recent study~\cite{nasiri2024vim4path} have explored pre-training Vim models with DINO for pathology images of various resolutions, and another recent work on autoregressive visual pretraining on ImageNet~\cite{ren2024autoregressive} investigates grouping $\frac{H}{16}\times \frac{W}{16}$ patches as prediction units, representing the first exploration of autoregressive pretraining with Mamba architectures. However, SSL strategies for Mamba models remain nascent. For this reason, we also explore applying DINO to pre-train Vim on GSV data. Experimental details are provided in the following paragraph.

\begin{table*}[t]
  \centering
  {%
  \setlength{\tabcolsep}{5pt}      
  \setlength{\extrarowheight}{0pt} 
  \begin{tabular}{cccccccccc}
    \toprule
    Model & Epochs & LR & Min LR & LR Warmup & Weight decay &
    Teacher Temp & Warmup Teacher Temp &
    Batch Size/GPU & \#GPUs \\
    \midrule
    ViT-S/16 & 100 & $5\times10^{-4}$ & $1\times10^{-6}$ & 10 & 0.04 & 0.04 & 0 & 64 & 2 \\
    \specialrule{.03em}{0.12ex}{0.12ex}
    Vim-S/16 & 100 & $2\times10^{-5}$ & $1\times10^{-6}$ & 10 & 0.04 & 0.04 & 0 & 32 & 4 \\
    \bottomrule
  \end{tabular}}
  \caption{Hyperparameters used for DINO post-training on 1 million GSV images.
“LR warmup” indicates the number of epochs for linear learning rate warmup, and “warmup teacher temp” indicates the number of epochs for teacher temperature warmup.}
  \label{tab:DINO round1 hyperparams}
\end{table*}

\subsection{Training Setting}\label{sec:DINO Training setting}
To balance model capacity with the computational cost of large-scale training, we focus on small models for post-training on unlabeled data. Specifically, we select ViT-S/16 pretrained on ImageNet in an unsupervised manner using DINO (21M parameters) and Vim-S pretrained on ImageNet-1K (26M parameters), and further post-train both on 1 million GSV images. Since training starts from ImageNet-pretrained weights, we refer to this unsupervised training process as post-training, as illustrated in Figure \ref{fig:training framework}. The 1M images are randomly sampled from our dataset without filtering, deduplication, or weighting. The necessity of data curation is discussed later in Section~\ref{sec:model evaluation}.

We adopt the head output dimensionality(65,536) from DINO. For both models, we use 2 global crops and 8 or 10 local crops, with 8 local crops as the default. Optimization is performed with AdamW. Key hyperparameters are reported in Table \ref{tab:DINO round1 hyperparams}, while the remaining settings follow the original DINO defaults. The learning rate (LR) in Table \ref{tab:DINO round1 hyperparams} corresponds to the LR at the end of the linear warmup, and the minimum LR represents the target LR at the end of optimization. A cosine LR scheduler is used, and DINO also applies a cosine schedule for weight decay, gradually increasing the decay to improve performance. Teacher temperature controls the smoothness of the teacher’s predicted distribution. DINO uses a warmup for teacher temperature, as a high temperature at the start can make training unstable. For ViT-S/16, we set the teacher temperature warmup epochs to 0, as the training was stable without warmup.

Vim-S is more unstable under DINO. In initial experiments with an LR of $5\times10^{-4}$ and a minimum LR of $1\times10^{-6}$ for the first 6 epochs, training diverged due to NaN losses. Several combinations of LR, minimum LR, and weight decay were attempted, but instability persisted. We then adopted the hyperparameters listed in Table \ref{tab:DINO round1 hyperparams} and resumed training from epoch 6 to 100. Teacher temperature warmup may also affect stability, but we leave a more systematic investigation of this for future work. We also explore scaling both model size and training data to medium-scale settings using 5 million images. However, training at this scale proves challenging; therefore, we report experimental results only for small models. The observed training instabilities are discussed in Appendix~\ref{apdx:medium-sized models}.

All post-training and evaluation/fine-tuning experiments in this work were conducted on NVIDIA L40S GPUs (48 GB) and NVIDIA RTX A6000 GPUs (48 GB). Training ViT-S/16 takes slightly over one hour per epoch on 1M images using 2×48GB GPUs, whereas training Vim-S/16 takes less than 2.5 hours per epoch using 4×48GB GPUs.

\subsection{Data Augmentation}
\begin{figure}[t] 
  \centering
  \includegraphics[width=0.5\textwidth]{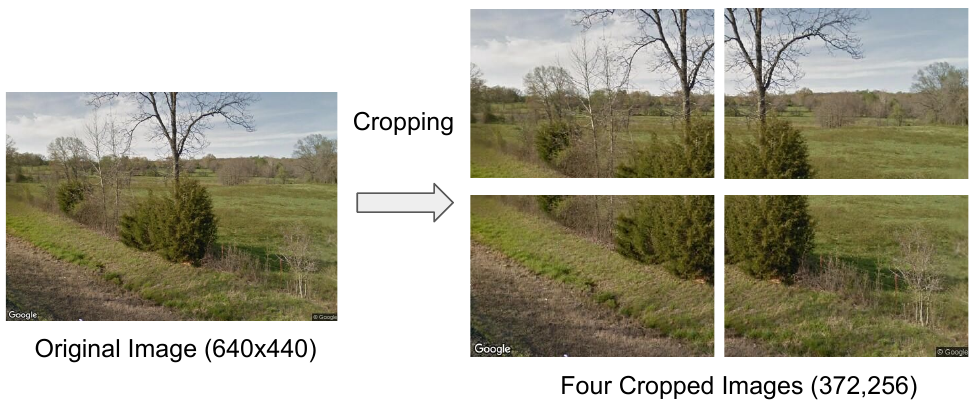}
  \caption{Overlapping crops for data augmentation used in DINO training.}
  \label{fig:DINOcropping}
\end{figure}

DINO was originally proposed for training on ImageNet, where image sizes and aspect ratios vary—for example, 200×200 pixels or 567×378 pixels. In contrast, all GSV images share the same size of $640 \times 440$. For the training described above, we adopt the same data augmentation pipeline as DINO on ImageNet. This includes a RandomResizedCropAndInterpolation with size $224 \times 224$ for global crops and $96 \times 96$ for local crops, along with random horizontal flipping, color jitter, Gaussian blur, and normalization.

For global crops, we by default select a crop area (scale) from 0.4 to 1.0 of the original image and randomly choose an aspect ratio between 0.75 and 1.333, then resize the crop to $224 \times 224$, as in ImageNet. Local crops use a default scale range of 0.05 to 0.4. We also experimented with other scale factors for different models. Since such cropping can change the aspect ratio of the input, we also propose a method to preserve the original aspect ratio: we crop the $640 \times 440$ images into four overlapping crops of size $372 \times 256$, as illustrated in Figure \ref{fig:DINOcropping}.

Using this new augmentation, we further train ViT-S starting from the weights obtained after the first round of training for an additional 50 epochs on the same 1M images. The number of local crops is increased to 10. We also increase the batch size to 128 per GPU and use 2 GPUs, as DINO has been reported to benefit from large batch sizes, e.g., 1024. The learning rate is set to $2.5\times10^{-4}$ with a minimum LR of $5\times10^{-7}$ and a weight decay of 0.5, as the training is resumed from the previous checkpoint. 

\section{Model Evaluation}\label{sec:model evaluation}
To evaluate representation quality after post-training, we assess downstream classification performance with a few thousand labeled samples per task under two settings: (i) fine-tuning the model using the available labeled data, and (ii) linear probing via a linear classifier trained on frozen backbone weights. Both settings are applied across all benchmark tasks. ViT-S and ViT-B use no labels during their pretraining and post-training stages prior to evaluation, and all models use labeled data at the evaluation stage. Class distributions are reported in Table \ref{tab:data split}. For each task, data are split in a stratified manner into training, validation, and test sets at a 75:10:15 ratio, except for Sidewalk, which uses a 70:15:15 split.

\subsection{Data Augmentation for Model Evaluation}\label{sec:data aug for model evaluation}
We use the same image augmentation strategy as in the post-training process, including a RandomResizedCropAndInterpolation with a crop scale between 0.08 and 1.0, a randomly sampled aspect ratio between 1.0 and 1.6, resizing to 224×224, followed by random horizontal flipping and normalization. The aspect ratio range was chosen empirically, as it performed better in most cases.

\begin{table}[!ht]
    \centering
    \begin{tabular}{cl|llll}
    \toprule
        \multicolumn{2}{c|}{Task/Class}                       & Train & Val & Test  & Total\\ \midrule
        \multicolumn{1}{c|}{\multirow{2}{*}{Streetlight}} & 0 & 11,844      & 1,579    &  2,369  & 15,792       \\ 
        \multicolumn{1}{c|}{}                             & 1 &  1,608     &  214   &   322    &  2,144   \\ \specialrule{.03em}{0.3ex}{0.3ex}
        \multicolumn{1}{c|}{\multirow{3}{*}{NSH}}         & 0 &   4,146    & 552    &  830    &   5,528 \\ 
        \multicolumn{1}{c|}{}                             & 1 &    5,430   &   724  &    1,087   & 7,241  \\ 
        \multicolumn{1}{c|}{}                             & 9 &    568   &    76 &       113  & 757\\ \specialrule{.03em}{0.3ex}{0.3ex}
        \multicolumn{1}{c|}{\multirow{2}{*}{Green30}}     & 0 &    3,122   &   416  & 624     &  4,162     \\ 
        \multicolumn{1}{c|}{}                             & 1 &     7,007  &    935 & 1,402     &  9,344     \\ \specialrule{.03em}{0.3ex}{0.3ex}
        \multicolumn{1}{c|}{\multirow{2}{*}{Sidewalk}}    & 0 &   9,641    &  2,066   &  2,066     &  13,773    \\ 
        \multicolumn{1}{c|}{}                             & 1 &   2,897    &   621  &       621    & 4,139 \\ 
    \bottomrule
    \end{tabular}
    \caption{Data split}
    \label{tab:data split}
\end{table}

\subsection{Fine-tuning}\label{fine-tuning}

\begin{table*}[t]
\renewcommand{\arraystretch}{1.25}
\centering
\resizebox{\textwidth}{!}{%
\begin{tabular}{c|c|c|c|c|c|c|c|c|c|c|c|c|c|c|c}
\toprule
\multirow{2}{*}{Model} &
\multirow{2}{*}{Size(M)} &
\multirow{2}{*}{\begin{tabular}[c]{@{}c@{}}ImageNet\\ Trained\end{tabular}} &
\multirow{2}{*}{\begin{tabular}[c]{@{}c@{}}GSV\\ Trained\end{tabular}} &
\multicolumn{3}{c|}{Streetlight} &
\multicolumn{3}{c|}{NSH} &
\multicolumn{3}{c|}{Green30} &
\multicolumn{3}{c}{Sidewalk}
\\ \cline{5-16}
 & & & &
Acc. & BAcc. & \begin{tabular}[c]{@{}c@{}}F1\\binary\end{tabular} &
Acc. & BAcc. & \begin{tabular}[c]{@{}c@{}}F1\\macro\end{tabular} &
Acc. & BAcc. & \begin{tabular}[c]{@{}c@{}}F1\\macro\end{tabular} &
Acc. & BAcc. & \begin{tabular}[c]{@{}c@{}}F1\\macro\end{tabular}
\\ \specialrule{.03em}{0.3ex}{0.3ex}

ResNet50 & 23.51 & \checkmark &  &
85.15 & 83.24 & 56.52 &
81.85 & 73.56 & 75.92 &
83.84 & 83.41 & 81.82 &
86.65 & 86.71 & 82.97
\\ \specialrule{.03em}{0.3ex}{0.3ex}

ViT-S & 21.67 & \checkmark & &
88.89 & 80.81 & 60.19 &
83.30 & 74.20 & 77.90 &
\textbf{86.62} & 82.15 & 83.56 &
90.32 & 86.05 & 86.29
\\ \specialrule{.03em}{0.3ex}{0.3ex}

ViT-S & 21.67 & \checkmark & \checkmark & 
87.66 & 81.45 & 58.71 & 
83.15 & 77.86 & 79.45 & 
86.03 & 83.37 & 83.53 & 
91.04 & 86.60 & 87.20 
\\ \specialrule{.03em}{0.3ex}{0.3ex}

Vim-S & 25.62 & \checkmark &  & 
\textbf{91.08} & 77.49 & 61.54 & 
84.04 & \underline{80.95} & 80.85 & 
86.18 & 82.68 & 83.41 & 
91.00 & 87.69 & 87.42 
\\ \specialrule{.03em}{0.3ex}{0.3ex}

Vim-S & 25.62 & \checkmark & \checkmark & 
89.48 & 82.49 & 62.52 & 
\underline{84.93} & \textbf{82.03} & \textbf{81.49} & 
85.88 & 84.24 & \underline{83.70} & 
\textbf{91.89} & \textbf{89.25} & \textbf{88.74} 
\\ \specialrule{.03em}{0.3ex}{0.3ex}

ViT-B & 85.81 & \checkmark &  & 
\underline{90.82} & 81.74 & \textbf{64.46} & 
\textbf{85.55} & 79.00 & \underline{80.99} & 
85.98 & 82.53 & 83.20 & 
\underline{91.74} & 85.81 & \underline{87.73} 
\\ \specialrule{.03em}{0.3ex}{0.3ex}

Vim-B & 96.83 & \checkmark &  & 
89.34 & 79.99 & 60.30 & 
82.59 & 78.69 & 78.59 & 
85.19 & \underline{84.37} & 83.19 & 
90.52 & 85.02 & 86.21 
\\ \specialrule{.03em}{0.3ex}{0.3ex}

Swin-S & 48.84 & \checkmark &  & 
88.63 & \underline{84.69} & 62.59 & 
82.84 & 76.21 & 78.53 & 
85.49 & 84.22 & 83.38 & 
88.84 & \underline{88.24} & 85.39 
\\ \specialrule{.03em}{0.3ex}{0.3ex}

Swin-B & 86.75 & \checkmark &  & 
87.96 & \textbf{87.12} & \underline{63.10} & 
83.19 & 74.74 & 77.69 & 
\underline{86.28} & \textbf{84.48} & \textbf{84.09} & 
89.46 & 87.89 & 85.91 
\\ \bottomrule
\end{tabular}%
}
\caption{Performance comparison on Streetlight, NSH, Green30, and Sidewalk datasets, fine-tuning the backbone.}
\label{tab: finetune performance on 4 tasks}
\end{table*}

To avoid overfitting, we fine-tune all models—including the backbone and a randomly initialized linear classifier—for 30 epochs across all tasks. Owing to the imbalanced class distributions shown in Table \ref{tab:data split}, we report balanced accuracy as the primary metric, alongside accuracy and F1 score. For highly imbalanced tasks such as Streetlight and Sidewalk, we use a balanced sampler to oversample minority classes. 

Among all models, the post-pretrained Vim-S achieves the best overall generalization across tasks, with Swin-B performing competitively. Notably, although Swin-B achieves the highest ImageNet(upstream) accuracy among all initial weights, it does not consistently perform best on our downstream tasks, despite having 61M more parameters than Vim-S. Swin-S and Vim-B achieve the second-best ImageNet (upstream) accuracy among all initial weights, but they do not generalize as well to GSV data. This illustrates that larger models do not necessarily outperform smaller ones without proper domain adaptation, even when they achieve higher upstream benchmark scores. 

Specifically, Vim-S and ViT-S pretrained on both ImageNet and GSV outperforms Vim-S and ViT-S pretrained only on ImageNet on all tasks except observing a slight performance drop on Streetlight, demonstrating the effectiveness of our unsupervised post-training. The accuracy drop of Vim-S on Streetlight after post-pretraining can be attributed to the use of balanced sampling: while it improves balanced accuracy, it may slightly reduce accuracy for majority classes.
We later examine the Streetlight performance drop of ViT-S using attention map visualizations, discussed in Section~\ref{sec:attention map}.


The parameter efficiency and inherent local inductive bias of Mamba-based models suggest opportunities for developing small yet effective architectures ideal for resource-constrained environments, such as Vim-S, consistent with recent findings\cite{xu2024visual}. However, scaling Mamba to larger network sizes remains challenging. In our experiments, scaling to Vim-B and training on 5M images leads to training collapse under DINO regardless of learning rate. Techniques such as gradient clipping and skipping samples that produce NaN losses offer little improvement. Prior studies also report instability in large Mamba variants\cite{patro2025mamba}, often caused by vanishing or exploding gradientss\cite{patro2024simba}, which can degrade performance or cause training failure. Consequently, most visual Mamba models remain at base or smaller scales, limiting their overall performance\cite{xu2024visual}.

By comparing checkpoints across two rounds of post-training, we propose three conjectures regarding the performance behavior of our trained models:
\begin{itemize}
\item \textbf{Longer unsupervised training leads to over-generalization.}
Prolonged unsupervised training can produce overly general representations, causing models to emphasize majority visual patterns—particularly those from remote areas overrepresented in our 1M dataset—which in turn lowers balanced accuracy on downstream tasks. This highlights both the strengths and limitations of using randomly sampled, uncurated, and unlabeled data for unsupervised training, and underscores the importance of data curation strategies, such as removing redundancy (including semantic redundancy), accounting for sampling bias rather than relying on random sampling, and applying sample-level reweighting when appropriate. 

This observation is consistent with findings in LLM training, where data quality often outweighs architectural improvements. Results in Table \ref{tab: finetune performance on 4 tasks} further show that solely oversampling minority classes is insufficient. Recent work has explored dynamically evaluating and weighting individual samples \cite{ye2024data}. Finally, the limited performance gains from post-training may also be attributed to the constraints of DINOv1’s memorization-based teacher.

\item \textbf{Capacity saturation in small models}
ViT-S and Vim-S (22M and 26M parameters) exhibit decreased performance when training is continued on the same 1M dataset in round 2, suggesting that small models may lack the capacity to continue benefiting from longer unsupervised training. This highlights the importance of selecting an appropriate model size that can effectively scale with dataset size while balancing computational cost.

\item \textbf{Necessity of data curation.}
Simply increasing the training set size not only slows training and increases computational and memory demands, but also introduces greater instability and leads to overfitting on small downstream tasks (see Appendix~\ref{apdx:medium-sized models}). This further underscores the importance of careful data curation during both pre-training and post-training, as incorporating more randomly sampled data does not necessarily improve downstream performance.

Recent deep learning research has elevated data curation from a tedious “data cleaning” task to a central scientific and engineering discipline, playing a decisive role in model performance, efficiency, security, and economic feasibility. High-quality, well-curated data is therefore essential for building more powerful, reliable, and larger-capacity models.
\end{itemize}

In summary, bridging the domain gap between pretraining data and downstream tasks requires carefully designed strategies, including improved data filtering, sampling, and learning techniques. Model architecture selection is likewise non-trivial and plays an essential role.

\begin{table*}[t]
\renewcommand{\arraystretch}{1.25}
\centering
\resizebox{\textwidth}{!}{%
\begin{tabular}{c|c|c|c|c|c|c|c|c|c|c|c|c|c|c}
\toprule
\multirow{2}{*}{Backbone} &
\multirow{2}{*}{\begin{tabular}[c]{@{}c@{}}ImageNet\\ Trained\end{tabular}} &
\multirow{2}{*}{\begin{tabular}[c]{@{}c@{}}GSV\\ Trained\end{tabular}} &
\multicolumn{3}{c|}{Streetlight} &
\multicolumn{3}{c|}{NSH} &
\multicolumn{3}{c|}{Green30} &
\multicolumn{3}{c}{Sidewalk}
\\ \cline{4-15}
 & & &
Acc. & BAcc. & \begin{tabular}[c]{@{}c@{}}F1\\binary\end{tabular} &
Acc. & BAcc. & \begin{tabular}[c]{@{}c@{}}F1\\macro\end{tabular} &
Acc. & BAcc. & \begin{tabular}[c]{@{}c@{}}F1\\macro\end{tabular} &
Acc. & BAcc. & \begin{tabular}[c]{@{}c@{}}F1\\macro\end{tabular}
\\ \specialrule{.03em}{0.3ex}{0.3ex}

ResNet50 & \checkmark &  &
73.02 & 79.44&43.81 &
 66.67&45.44 &44.23 &
 80.36&80.34 & 78.28&
71.89 & 76.81& 68.65
\\ \specialrule{.03em}{0.3ex}{0.3ex}

ViT-S & \checkmark & &
 77.04& 78.64& 45.69&
68.29 &\underline{51.92} &\textbf{53.26} &
77.74 &69.51 & 71.03&
80.38 &69.83 & 70.86
\\ \specialrule{.03em}{0.3ex}{0.3ex}

ViT-S & \checkmark & \checkmark &
78.93& 75.42& 44.57&
62.92& 48.49&49.42 &
74.63& 67.26& 68.18&
 77.48&68.32 & 68.32
\\ \specialrule{.03em}{0.3ex}{0.3ex}

Vim-S & \checkmark &  & 
76.33 & 80.38&46.43 &
 71.20 & 51.03&49.35 &
81.84 & 80.07& 79.24&
81.10 & 69.85& 71.28
\\ \specialrule{.03em}{0.3ex}{0.3ex}

Vim-S & \checkmark & \checkmark & 
\textbf{80.53} &79.95 & \textbf{49.32}&
 68.00& 48.55&48.04 &
 81.44& 79.30&78.67 &
 \underline{81.99}& 71.18& 72.68
\\ \specialrule{.03em}{0.3ex}{0.3ex}

ViT-B & \checkmark &  & 
77.33 & 78.94 & 46.11 &
68.10 & 48.40 & 48.26 &
78.78 & 77.95&76.34 &
79.86 & 68.14& 69.42
\\ \specialrule{.03em}{0.3ex}{0.3ex}

Vim-B & \checkmark &  & 
75.85 & \underline{80.39} & 46.10 & 
\underline{71.89} & 50.64 & 49.11 & 
 \textbf{85.04}&\textbf{84.35} &  \textbf{83.06}&
\textbf{83.06} & 73.45& \textbf{74.75}
\\ \specialrule{.03em}{0.3ex}{0.3ex}

Swin-S & \checkmark &  & 
 77.37&80.30 & 47.09&
 71.30&51.37 & 49.44&
\underline{83.32}&\underline{81.54} & \underline{80.84}&
78.54 &\underline{81.02} & \underline{74.70}
\\ \specialrule{.03em}{0.3ex}{0.3ex}

Swin-B & \checkmark &  & 
 \underline{79.23}& \textbf{81.22}&\underline{49.14} &
 \textbf{73.87}& \textbf{52.53}& \underline{50.59}&
 80.85& 80.25& 78.61&
77.80 &\textbf{81.21} & 74.21
\\ \bottomrule
\end{tabular}%
}
\caption{Linear probing on Streetlight, NSH, Green30, and Sidewalk datasets. All backbones are frozen.}
\label{tab:freeze backbone performance on 4 tasks}
\end{table*}

\begin{figure*}[!htbp]  
  \centering
  \includegraphics[width=\textwidth]{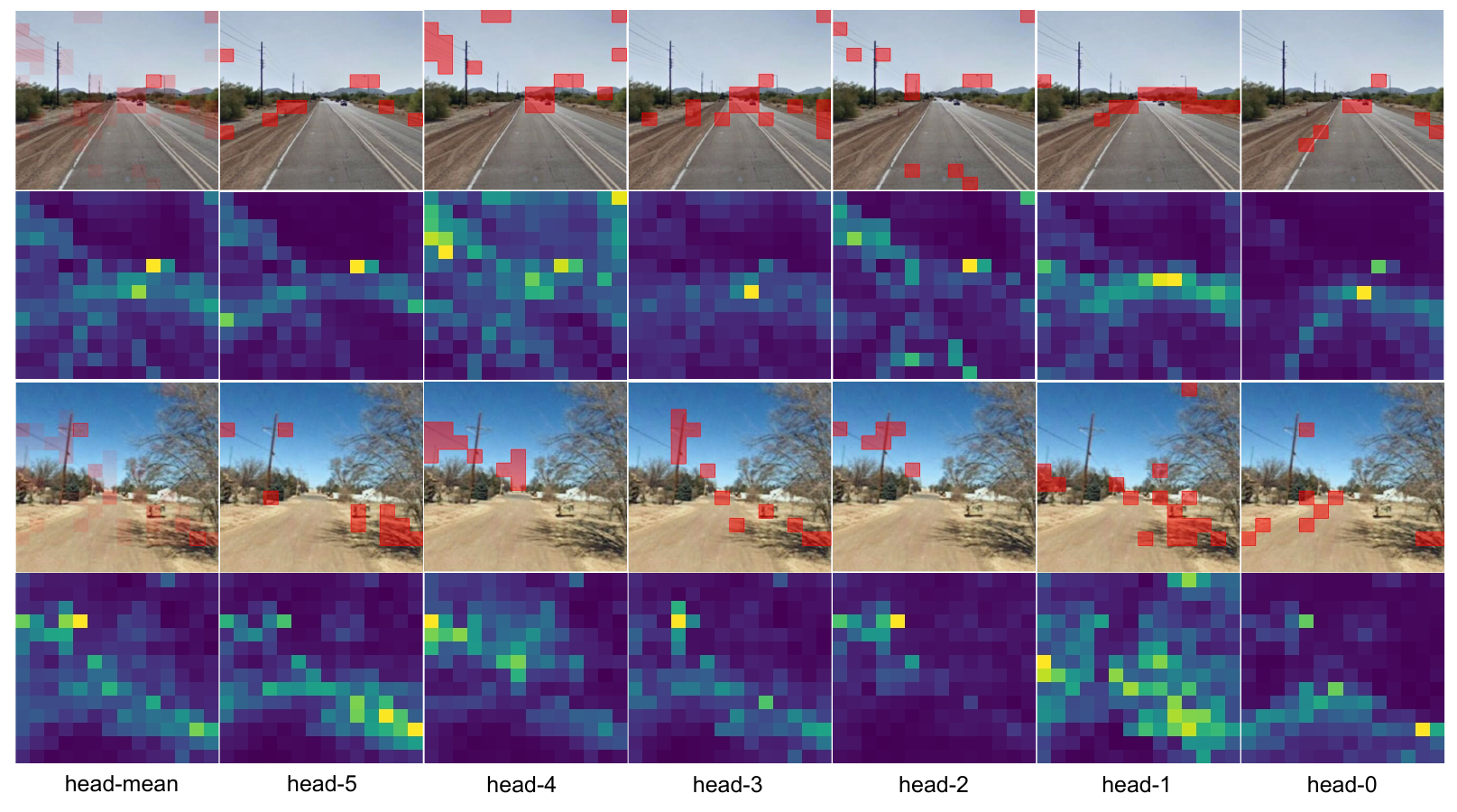}
  \caption{A positive(top) sample and a negative(bottom) sample randomly selected from the test set of streetlight classification. A ViT-S unsupervisedly pretrained on ImageNet and GSV data is used. Top: Ground truth: 1; Prediction: 1; Bottom: Ground truth: 0; Prediction: 0.}
  \label{fig:ViT-S(IN,GSV) attention map on streetlight}
\end{figure*}

\subsection{Linear Probing}\label{linear probe}
To further evaluate the representation quality of backbones, we conduct linear probing (also known as linear evaluation \cite{chen2020simple}), in which a linear classification layer is trained on labeled downstream data across four tasks while the backbone remains frozen. In this widely used protocol, test accuracy assesses representation quality by measuring the linear separability and informativeness of the fixed learned features. As shown in Table~\ref{tab:freeze backbone performance on 4 tasks}, GSV-post-trained backbones do not consistently outperform models pretrained solely on ImageNet. In several cases (e.g., ViT-S across all tasks and Vim-S on NSH and Green30), additional GSV post-training even leads to performance degradation. This is notable because domain-adaptive pretraining is typically expected to enhance downstream transferability. We hypothesize several possible reasons for this behavior:

\begin{itemize}
\item \textbf{Domain shift.}
Although all downstream tasks (Streetlight, NSH, Green30, Sidewalk) involve urban-scene understanding, their visual statistics differ substantially from raw GSV imagery. For example, Green30 contains vegetation-dominant scenes with lighter textures, while Sidewalk focuses heavily on ground-plane patterns. By contrast, raw GSV frames often contain wide fields-of-view, high clutter, inconsistent resolution, irrelevant backgrounds, and varied lighting or lens artifacts. Post-training on GSV may cause the model to overfit these statistics, which do not fully align with the downstream domains. Consequently, ImageNet-pretrained models can sometimes be “straighter” representations, whereas GSV post-training acts as a domain perturbation that is misaligned when the backbone remains frozen.

\item \textbf{A frozen backbone amplifies the impact of domain mismatch.}  
If the backbone were fine-tuned end-to-end, the model could correct the misleading biases introduced by raw GSV. However, in the linear probing setting, a domain-adapted (but misaligned) representation may perform less robustly across tasks than ImageNet features.

\item \textbf{Potential representation drift from extended unsupervised training.}  
ImageNet pretraining is curated and object-centric, whereas GSV data are uncurated and noisy. Prolonged post-training on GSV may encourage the model to learn spurious background cues, geographic biases, or other non-semantic correlations, while frozen heads cannot compensate for these biases. In Table~\ref{tab:freeze backbone performance on 4 tasks}, results for ViT-S use the same 92-epoch checkpoint as in Table~\ref{tab: finetune performance on 4 tasks}. We also observed that checkpoints trained slightly longer perform worse, whereas earlier checkpoints perform better, supporting the hypothesis that longer unsupervised training can exacerbate representation drift.
\end{itemize}

\section{Attention Maps}\label{sec:attention map}
To investigate the performance drop of ViT-S post-trained on GSV on the Streetlight task in Table~\ref{tab:freeze backbone performance on 4 tasks}, we visualize the attention maps for two randomly selected Streetlight test samples(Figure~\ref{fig:ViT-S(IN,GSV) attention map on streetlight}). For clarity, we threshold the top 20\% of attention scores. In addition to correctly classified cases, we identify two typical failure modes, described below.

\begin{figure*}[!htbp]
  \centering
  \includegraphics[width=0.71\textwidth,keepaspectratio=false, height=0.29\textwidth]{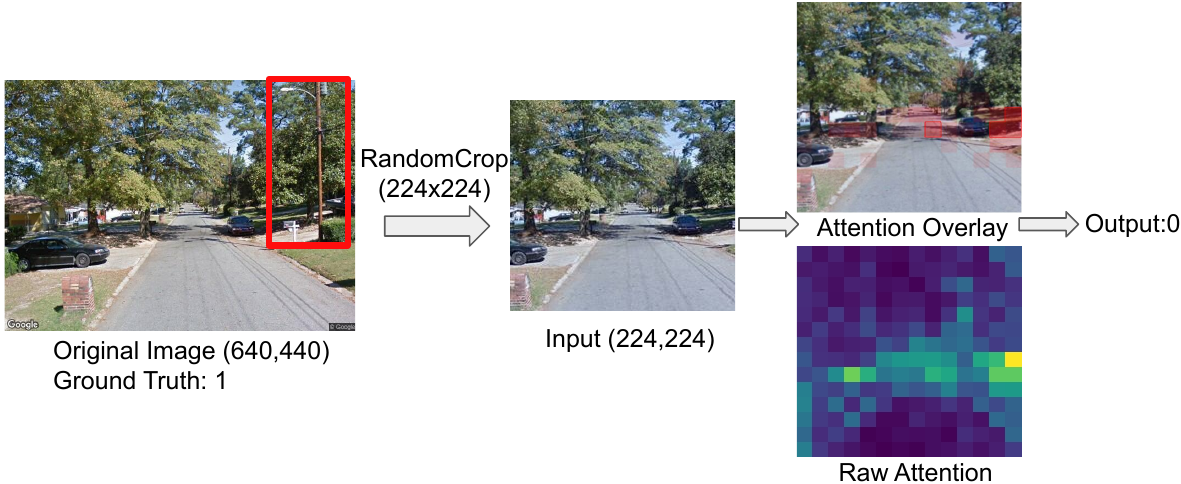}
  \caption{Failure case 1: information loss caused by input size limits in ViT models. Mean attention over all heads are visualized. The object is highlighted by a red rectangle in the left figure.}
  \label{fig:vit_s_streetlight_failurecase1}
\end{figure*}

\subsection{Failure Case: Information Loss from Input Size Constraints}
In Figure~\ref{fig:vit_s_streetlight_failurecase1}, an image containing a streetlight at the right boundary was labeled positive but predicted negative by ViT-S. ViT-based architectures rely on fixed square input sizes (commonly 224×224 for efficiency), which can lead to information loss when important objects fall outside the crop or are removed during preprocessing. We apply a cropping and voting strategy to our data, consistent with the second round of post-training (Figure~\ref{fig:DINOcropping}).


Here we remove distillation loss and mixup augmentation (used in Section~\ref{fine-tuning} and Section~\ref{linear probe}, adopted from Vim~\cite{zhu2024vision}). We keep the same data augmentations except for the addition of overlapping cropping. For each input image, we perform forward passes on four crops independently and determine the final prediction by selecting the maximum positive-class probability, reducing the risk of missing the target object. 

Experimental results show that the cropping and voting strategy partially alleviates the information loss observed in Failure Case 1 (Figure~\ref{fig:vit_s_streetlight_failurecase1}), but it cannot fully prevent misses. Based on Table~\ref{tab:FourCropVoting}, we conjecture that FourCropVoting increases true positive predictions(TP) by reducing missed detections, but at the cost of producing more false positives(FP) due to overconfident positive predictions. As negative samples dominate the dataset, this results in lower overall accuracy while improving balanced accuracy. The F1 score slightly decreases, likely due to reduced precision caused by the additional false positives. These findings indicate that more carefully designed solutions are needed when using ViT-based models that operate on fixed input sizes and cannot naturally scale to non-square or high-resolution images without specific architectural modifications.
\begin{table}[t]
    \centering
    \setlength{\abovecaptionskip}{3pt} 
    \begin{tabular}{cccc}
    \toprule
    {Data Processing} & {Acc.} & Bacc. & F1(binary) \\ \specialrule{.03em}{0.3ex}{0.3ex}
    {RandomCrop (224)} & 88.52 & 76.84 & 56.17 \\ \specialrule{.03em}{0.3ex}{0.3ex}
    {FourCropVoting}  & 85.17 & 82.32 & 55.91 \\ \bottomrule
    \end{tabular}
    \caption{Comparison of two data processing strategies based on fine-tuning ViT-S for streetlight classification.}
    \label{tab:FourCropVoting}
\end{table}

\subsection{Failure case: Different appearance}
As shown in Figure~\ref{fig:vit_s_streetlight_failurecase2}, our trained ViT-S sometimes fails to recognize streetlights whose appearance differs from those in correctly classified examples (Figure~\ref{fig:ViT-S(IN,GSV) attention map on streetlight}). This accounts for the observed ~1.5\% drop in binary F1 and ~0.6\% in balanced accuracy for the Streetlight classification task.
\begin{figure}[!htbp]
    \centering
    \includegraphics[width=0.75\linewidth]{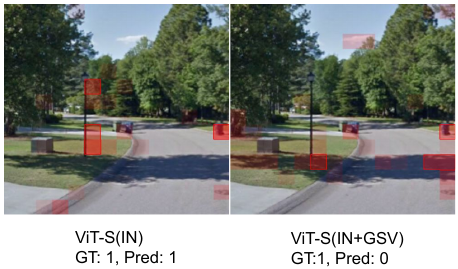}
    \caption{Failure case 2: Different appearance. Mean attention over all heads are visualized.}
    \label{fig:vit_s_streetlight_failurecase2}
\end{figure}

\section{Discussion}
Bridging the domain gap requires careful choices of model architecture, SSL strategy, and data sampling. Unsupervised post-training generally brings model representations closer to street-view semantics and provides a stronger initialization for downstream tasks. However, when applied to uncurated GSV data, it can introduce domain-specific biases.Without careful sampling, smaller models saturate quickly and learn less transferable features as the dataset scales, reducing downstream performance gains. Curated data combined with well-designed SSL techniques can therefore be more effective, particularly for Mamba architectures.

Architecture choice is critical. Stronger upstream performance does not necessarily translate to better downstream results, as generalization is also architecture-dependent. For example, Swin-B underperforms Swin-S and Vim-B in linear probing. Our experiments further show that the benefits of post-training vary substantially across architectures. Overall, Swin Transformers remain reliable among ViT-based models due to their built-in spatial priors, while Mamba-based architectures show strong potential. The Vim architecture appears relatively robust to domain shift—Vim-S remains competitive with larger models after GSV post-training—but exhibits notable training instabilities when scaling to larger model sizes or datasets. Developing robust training strategies for Mamba-based models remains an important direction for future work.

Finally, we find that although unsupervised models are often assumed to learn more generalizable representations, supervised pretraining provides stronger initial performance, benefiting downstream tasks.

\clearpage 
{\small
\bibliographystyle{ieee_fullname}
\bibliography{egbib}
}

\appendix

{

\subsection{Scaling models}\label{apdx:medium-sized models}
After two rounds of unsupervised training on GSV, we explore scaling both the model size and the training data to medium-scale models using 5 million images. We randomly sampled 5M GSV images and selected the Vision Transformer (ViT-Base/8) with patch size 8 and 85M parameters, pretrained unsupervisedly on ImageNet using DINO. ViT-Base/8 achieves approximately 3\% higher linear accuracy (80.1\%) on ImageNet compared to ViT-S/16 (77\%) used in the first round, as shown in Table \ref{tab:model variants}. The improved performance likely benefits from both the larger model size and finer patch resolution. We also scale Vim to Vision Mamba (Vim-Base), pretrained on ImageNet-1K in a supervised manner with 98M parameters. For Vim-Base, the global crop scale is set between 0.25 and 1.0, and the local crop scale is set between 0.05 and 0.25, the same as for ViT-B. Other training hyperparameters are listed in Table \ref{tab:medium models DINO round1 hyperparams}.

\begin{table}[h]
    \centering
    \begin{tabular}{cccc}
    \toprule
    LR & Min LR & LR Warmup & Weight Decay \\
    \midrule
    $5\times10^{-4}$ & $2\times10^{-6}$ & 10 & 0.4 \\
    $2\times10^{-6}$ & $1\times10^{-6}$ & 10 & $1\times10^{-8}$ \\
    \specialrule{.03em}{0.3ex}{0.3ex}
    Teacher Temp &
    \makecell{Warmup\\Teacher Temp} &
    \makecell{Batch Size\\ / GPU} &
    \#GPUs \\
    \specialrule{.03em}{0.3ex}{0.3ex}
    0.07 & 50 & 6  & 4 \\
    0.04 & 0  & 32 & 4 \\
    \bottomrule
    \end{tabular}
    \caption{Hyperparameters used for medium-sized models with DINO post-training on 5 million GSV images. Rows correspond to ViT-B/8 (top) and Vim-B/16 (bottom). “LR Warmup” indicates the number of epochs for linear learning rate warmup, and “Warmup Teacher Temp” indicates the number of epochs for teacher temperature warmup.}
    \label{tab:medium models DINO round1 hyperparams}
\end{table}

Training memory requirements grow substantially with model size and finer patch resolution. Using 4 GPUs, one epoch of ViT-Base/8 on 5M images requires around 36 hours. Training proved unstable: the model crashed after 6 epochs, highlighting the challenges and costs of scaling both model and dataset size. We provide a brief discussion on evaluating the crashed models in Section~\ref{sec:model evaluation}.

Similar to Vim-S, Vim-Base is more numerically unstable than ViT under unsupervised DINO training. Interestingly, we observed that as the model scales, Vim is more computationally efficient than ViT, demonstrating its efficiency as an SSM-based architecture (see Section \ref{section:SSM}), whereas ViT tends to be more memory and computation efficient for smaller models. For Vim-B/16, training on 5M images with 4 GPUs requires around 10 hours per epoch.

However, training Vim-B at this scale is challenging. The model frequently crashes, stopping at the end of the first epoch due to NaN or Inf values in the student outputs. Adjusting the learning rate and weight decay schedulers provided limited improvement. We hypothesize that this instability stems from an incompatibility between the Vim architecture and the DINO training procedure. 

\subsection{Data Augmentation for Model Evaluation}
    Besides the image augmentation described in Section~\ref{sec:data aug for model evaluation}, we also apply repeated augmentation following Vim~\cite{zhu2024vision}, where the sampler repeatedly selects the same index while the dataset applies random augmentations, enabling each GPU to receive different augmented views. Without repeated augmentation, each sample would be seen only once per epoch per GPU. This approach provides more diverse augmentations over time and helps reduce overfitting.

\subsection{Additional Experiments in Fine-tuning}\label{apdx:additional observation}
    \subsubsection{Performance Drop in Longer-Trained Models}\label{apdx:round2 checkpoints}
    Additional observations from two rounds of post-training include: (i) the ViT-S model trained for 92 epochs in round 1 outperforms its counterparts trained for 100–150 epochs in round 2 using the cropping strategy shown in Fig.~\ref{fig:DINOcropping}; and (ii) the Vim-S model trained for only 5 epochs before crashing in round 1 performs better than versions trained for 20–100 epochs in round 2 with a smaller learning rate, which highlights the over-
    generalization. Finally, although ViT-B pretrained on ImageNet-22k with DINO achieves the best performance on one indicator (presence of streetlight), it generalizes poorly on others, further reinforcing the domain gap.

    \subsubsection{Supervised Training}\label{apdx:combined supervised training}
    To make full use of the small labeled portion of our data, we construct a multi-label, multi-class task to further fine-tune both ImageNet-pretrained models and GSV-post-trained models. Because only some labels are available for each image, unassigned labels are treated as unknown, and we train using only the provided labels as targets through a masked cross-entropy loss. We also experimented with transferring a checkpoint fine-tuned on one task (e.g., Streetlight) to another task (e.g., NSH), but supervised training offered limited improvement and exhibited a strong tendency to overfit. This further underscores the necessity of substantial pre-training or post-training to obtain models that generalize well.
}
\end{document}